\documentclass[runningheads]{llncs}

\usepackage{authblk}
\usepackage{graphicx}
\usepackage{amsmath}
\usepackage{amssymb}
\usepackage{dsfont}
\usepackage{booktabs}
\usepackage{algorithm}
\usepackage{algpseudocode}
\usepackage{algcompatible}
\usepackage{multirow}
\usepackage{makecell}
\usepackage{xcolor}
\usepackage{colortbl}
\usepackage{graphicx,caption}

\usepackage{wrapfig,lipsum}

\usepackage[colorlinks=true, allcolors=blue]{hyperref}

\usepackage[misc]{ifsym}

\begin{document}

\title{Knowledge Distillation to Ensemble Global and Interpretable Prototype-based Mammogram Classification Models}

\titlerunning{BRAIxProtoPNet++ for Interpretable Mammogram Classification}

\authorrunning{C. Wang et al.}

\author{Chong Wang\textsuperscript{1(\Letter)}, Yuanhong Chen\textsuperscript{1}, Yuyuan Liu\textsuperscript{1}, Yu Tian\textsuperscript{1}, Fengbei Liu\textsuperscript{1}, Davis J. McCarthy\textsuperscript{2}, Michael Elliott\textsuperscript{2}, Helen Frazer\textsuperscript{3}, Gustavo Carneiro\textsuperscript{1}}

\institute{
 \textsuperscript{1} Australian Institute for Machine Learning, The University of Adelaide, Adelaide, Australia \\
 \{chong.wang\}@adelaide.edu.au\\
 \textsuperscript{2} St Vincent's Institute of Medical Research, Melbourne, Australia \\
 \textsuperscript{3} St Vincent's Hospital Melbourne, Melbourne, Australia\\
}

%
\maketitle              
\begin{abstract}
State-of-the-art (SOTA) deep learning mammogram classifiers, trained with weakly-labelled images,
often rely on global models that produce predictions with limited interpretability, which is a key barrier to their successful translation into clinical practice.
On the other hand, prototype-based models improve interpretability by associating predictions with training image prototypes, but they are less accurate than global models and their prototypes tend to have poor diversity.
We address these two issues with the proposal of BRAIxProtoPNet++, which adds interpretability to a global model by ensembling it with a prototype-based model. BRAIxProtoPNet++ distills the knowledge of the global model when training the prototype-based model with the goal of increasing the classification accuracy of the ensemble.
Moreover, we propose an approach to increase prototype diversity by guaranteeing that all prototypes are  associated with different training images.
Experiments on weakly-labelled private and public datasets show that BRAIxProtoPNet++ has higher classification accuracy than SOTA global and prototype-based models. Using lesion localisation to assess model interpretability, we show BRAIxProtoPNet++ is more effective than other prototype-based models and post-hoc explanation of global models. Finally, we show that the diversity of the prototypes learned by BRAIxProtoPNet++ is superior to SOTA prototype-based approaches.

\end{abstract}

\section{Introduction}

Deep learning models~\cite{krizhevsky2012imagenet,lecun2015deep} have shown promising performance in many medical imaging applications (e.g.,
mammography~\cite{shen2021interpretable},
radiology~\cite{hermoza2020region}, diagnostics~\cite{kleppe2021designing}, ophthalmology~\cite{fang2019attention,fang2017automatic}), even achieving accuracy as high as human radiologists~\cite{wu2019deep}.
Regardless of the encouraging performance
of these models trained with weakly-labelled images, 
the limited interpretability of their predictions remains a barrier to their successful translation into clinical practice~\cite{rudin2019stop}.
Recently, some studies use post-hoc explanations (e.g., Grad-CAM~\cite{selvaraju2017grad}) to highlight image regions associated with model predictions. However, such highlighted classification-relevant image regions are often not reliable and insufficient for interpretability~\cite{rudin2019stop}.

There is a growing interest in the development of effective interpretable methods for medical image classifiers trained with weakly-labelled images. 
Khakzar et al.~\cite{khakzar2019learning} train a chest X-ray classifier with perturbed adversarial samples to form more reliable class activation maps (CAM)~\cite{selvaraju2017grad}.
Ilanchezian et al.~\cite{ilanchezian2021interpretable} introduce BagNets for interpretable gender classification in retinal fundus images, which can reveal how local image evidence is integrated into global image decisions. 
Chen et al.~\cite{chen2019looks} present ProtoPNet that learns a set of class-specific prototypes, and a test image is classified by evaluating its similarity to these prototypes, which provides a unique understanding of the inner workings of the model.
Furthermore, XProtoNet~\cite{kim2021xprotonet} learns prototypes with disease occurrence maps for interpretable chest X-ray classification.
To make the model focus on clinically interpretable features, additional signals (e.g., nuclei and fat droplets) are provided to supervise the attention map of a global biopsy image classifier~\cite{yin2021focusing}. 
In general, the above approaches are either based on poorly interpretable post-hoc explanations from highly accurate global classifiers or achieve good interpretability from less accurate local (e.g., prototype-based) classifiers.

In this paper, we present BRAIxProtoPNet++, a novel, accurate, and interpretable mammogram classification model. 
BRAIxProtoPNet++ ensembles a highly accurate global classification model  with an interpretable ProtoPNet model,
with the goal of achieving better classification accuracy than both  global and ProtoPNet models and satisfactory interpretability.
This goal is achieved by distilling the knowledge from the global model when training the ProtoPNet.
Furthermore, BRAIxProtoPNet++ increases the prototype diversity from ProtoPNet, with a new prototype selection strategy that guarantees that the prototypes are associated with a diverse set of training images.
To summarise, \textbf{our contributions are}: 1) a new approach to add interpretability to accurate global mammogram classifiers; 2) a new global and prototype-based ensemble model, named BRAIxProtoPNet++, trained with knowledge distillation to enable effective interpretability and higher classification accuracy than both the global and prototype-based models; and 3) improved prototype diversity of BRAIxProtoPNet++ compared to previous  prototype-based models~\cite{chen2019looks}.
Experimental results on weakly-supervised private and public datasets~\cite{cai2019breast} show that BRAIxProtoPNet++ improves classification accuracy and exhibits promising interpretability results compared to existing non-interpretable global classifiers and recently proposed interpretable models.

\section{Proposed Method}

In this section, we introduce our proposed method that relies on the weakly-labelled dataset $\mathcal{D} = \{ (\mathbf{x},\mathbf{y})_i \}_{i=1}^{|\mathcal{D}|}$, where $\mathbf{x} \in \mathbb{R}^{H \times W}$ represents the mammogram of size $H \times W$, and $\mathbf{y} \in \{0,1\}^2$ denotes a one-hot representation of the image label (e.g., cancer versus non-cancer).

\subsection{BRAIxProtoPNet++}

\begin{figure}[t!]
    \vspace{-5pt}
    \centering
    \includegraphics[width=\textwidth]{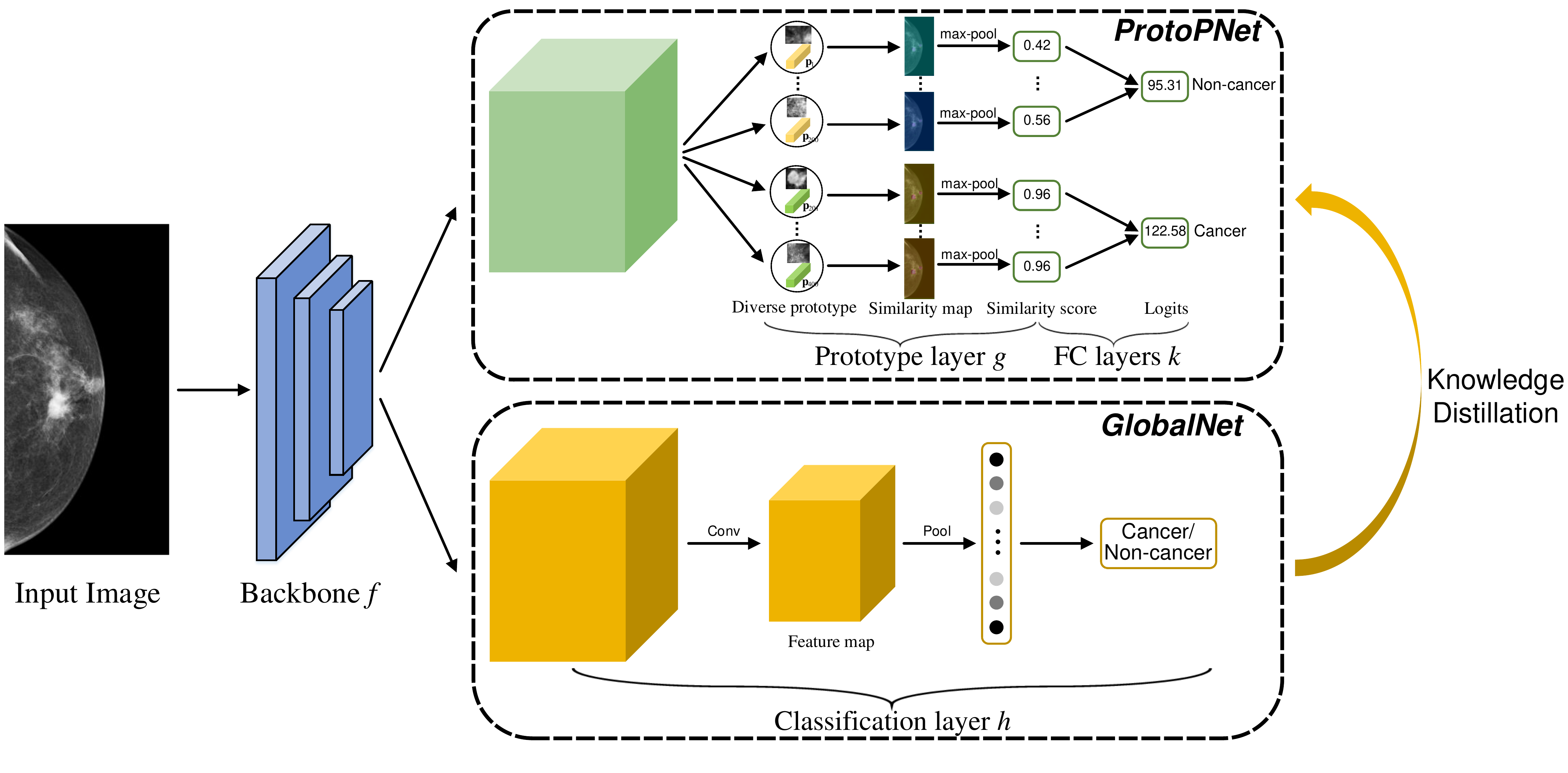}
    \vspace{-18pt}
    \caption{The architecture of our proposed BRAIxProtoPNet++, consisting of a shared CNN backbone, a global classifier GlobalNet, and an interpretable ProtoPNet that distils the knowledge from GlobalNet, to form an accurate ensemble classifier, and maximises the prototype diversity during training.
    }
    \label{fig:BRAIxProtoPNet++}
    \vspace{-12pt}
\end{figure}

Our proposed BRAIxProtoPNet++ with knowledge distillation and diverse prototypes, depicted in Fig.~\ref{fig:BRAIxProtoPNet++}, takes an accurate global image classifier (GlobalNet), trained with the weakly-labelled training set  $\mathcal{D}$, and integrates it with the interpretable ProtoPNet model.
ProtoPNet achieves classification by comparing local image parts with learned prototypes, which tends not to be as accurate as the holistic classification by GlobalNet.
This happens because even if local abnormalities are crucial for identifying breast cancer, information from the whole mammogram (e.g., lesions spreading in different spatial locations, or contrast between healthy and abnormal regions) may help to reach accurate classification.
Hence, to improve ProtoPNet's accuracy, we propose to distill the knowledge of GlobalNet to train ProtoPNet, using our new knowledge distillation (KD) loss function. 
Another limitation of ProtoPNet is the poor diversity of the learned prototypes that can negatively impact classification accuracy and model interpretability. We address this issue with a new prototype selection approach to increase the prototype diversity of the original ProtoPNet.

\subsubsection{Model} BRAIxProtoPNet++ comprises a CNN backbone (e.g., DenseNet \cite{huang2017densely} or EfficientNet \cite{tan2019efficientnet}) represented by $\mathbf{X}=f_{\theta_{f}}(\mathbf{x})$, where $\mathbf{X} \in \mathbb{R}^{\frac{H}{32} \times \frac{W}{32} \times D}$, and $\theta_{f}$ denotes the backbone parameters. The classification layer of the GlobalNet is denoted by $\tilde{\mathbf{y}}^{G} =h_{\theta_{h}}(\mathbf{X})$, where $\tilde{\mathbf{y}}^{G} \in [0,1]^2$ is the model prediction. 
The ProtoPNet is denoted by $\tilde{\mathbf{y}}^{L}=k_{\theta_{k}}(g_{\theta_{g}}(\mathbf{X}))$, where $\tilde{\mathbf{y}}^{L} \in [0,1]^2$ is the model prediction,
$g_{\theta_{g}}(\cdot)$ represents the prototype layer, and $k_{\theta_{k}}(\cdot)$ the  fully connected (FC) layers.
The prototype layer has $M$ learnable class-representative prototypes $\mathcal{P}= \{ \mathbf{p}_m\}_{m=1}^{M}$, with $M/2$ prototypes for each class and $\mathbf{p}_m \in \mathbb{R}^{D}$, 
which are used to form similarity maps 
$\mathbf{S}_m(h,w) = e^{\frac{-||\mathbf{X}(h,w)-\mathbf{p}_{m}||_2^2}{T}}$,
where $h \in \{1, ..., \frac{H}{32}\}$ and $w \in \{1, ..., \frac{W}{32}\}$ denote spatial indexes in similarity maps, and $T$ is a temperature factor.
The prototype layer $g_{\theta_{g}}(\cdot)$ outputs $M$ similarity scores from max-pooling $\Big\{\underset{h,w}\max\mathbf{S}_m(h,w)\Big\}_{m=1}^M$, which are fed to  $k_{\theta_{k}}(\cdot)$ to obtain classification result. 

\subsubsection{Training} BRAIxProtoPNet++ is trained by minimising the following objective:
\begin{equation}
\begin{split}
    \ell_{PPN++}(\mathcal{D},\theta_f,\theta_g,\theta_k,\theta_h,\mathcal{P}) =  &   \ell_{PPN}(\mathcal{D},\theta_f,\theta_g,\theta_k,\mathcal{P})  + \\
    & \alpha \ell_{CEG}(\mathcal{D},\theta_f,\theta_h) +  \beta \ell_{KD}(\mathcal{D},\theta_f,\theta_g,\theta_k,\mathcal{P}),
\end{split}
\label{eq:ell_PPN_PP}
\end{equation}
where $\alpha$ and $\beta$ are weighting hyper-parameters, 
$\ell_{PPN}(\mathcal{D},\theta_f,\theta_g,\theta_k,\mathcal{P})$ is the ProtoPNet loss defined in~\eqref{eq:ell_PPN}, $\ell_{CEG}(\mathcal{D},\theta_f,\theta_h)$ represents the cross-entropy loss to train $\theta_{f}$ and $\theta_{h}$ for  GlobalNet using label $\mathbf{y}$ and prediction $\tilde{\mathbf{y}}^{G}$, and
\begin{equation}
    \ell_{KD}(\mathcal{D},\theta_f,\theta_g,\theta_k,\mathcal{P}) = \frac{1}{|\mathcal{D}|} \sum_{i=1}^{|\mathcal{D}|} \max(0, (\mathbf{y}_i)^{\top}(\tilde{\mathbf{y}}_i^{G}) -  (\mathbf{y}_i)^{\top}(\tilde{\mathbf{y}}_i^{L}) + \omega)
    \label{eq:ell_KD}
\end{equation}
is our proposed knowledge distillation (KD) loss, with $(\mathbf{y}_i)^{\top}(\tilde{\mathbf{y}}_i^{G})$ and $(\mathbf{y}_i)^{\top}(\tilde{\mathbf{y}}_i^{L})$ denoting the predicted probability of the labelled class from the GlobalNet and ProtoPNet, and $\omega > 0$ representing a pre-defined margin to control  ProtoPNet’s confidence gain.
Our novel KD loss in~\eqref{eq:ell_KD} is designed to distill the knowledge~\cite{bucilu2006model} from GlobalNet to ProtoPNet to increase the classification accuracy of ProtoPNet and enable a better ensemble classification using both models.

The ProtoPNet loss introduced in~\eqref{eq:ell_PPN_PP} is defined by: 
\begin{equation}
\begin{split}
    \ell_{PPN}(\mathcal{D},\theta_f,\theta_g,\theta_k,\mathcal{P}) = & \ell_{CEL}(\mathcal{D},\theta_f,\theta_g,\theta_k,\mathcal{P}) + \\ 
    & \lambda_1 \ell_{CT}(\mathcal{D},\theta_f,\theta_g,\mathcal{P}) + \lambda_2 \max(0, \gamma- \ell_{SP}(\mathcal{D},\theta_f,\theta_g,\mathcal{P})),
\end{split}
\label{eq:ell_PPN}
\end{equation}
where $\lambda_1$,  $\lambda_2$, and $\gamma$ denote  hyper-parameters,  $\ell_{CEL}(\mathcal{D},\theta_f,\theta_g,\theta_k,\mathcal{P})$ is the cross-entropy loss between the label $\mathbf{y}$ and the ProtoPNet output $\tilde{\mathbf{y}}^{L}$, and 
\begin{equation}
    \ell_{CT}(\mathcal{D},\theta_f,\theta_g,\mathcal{P}) = \frac{1}{|\mathcal{D}|} \sum_{i=1}^{|\mathcal{D}|}\min_{\mathbf{p}_m\in\mathcal{P}_{\mathbf{y}_i}} \min_{\mathbf{z} \in\mathbf{X}_i}||\mathbf{z}-\mathbf{p}_m||_2^2,
    \label{eq:ell_CT}
\end{equation}
\begin{equation}
    \ell_{SP}(\mathcal{D},\theta_f,\theta_g,\mathcal{P}) = \frac{1}{|\mathcal{D}|} \sum_{i=1}^{|\mathcal{D}|}\min_{\mathbf{p}_m\notin\mathcal{P}_{\mathbf{y}_i}} \min_{\mathbf{z} \in\mathbf{X}_i}||\mathbf{z}-\mathbf{p}_m||_2^2,
    \label{eq:ell_SP}
\end{equation}
where we abuse the notation to represent $\mathbf{z} \in \mathbb{R}^{D}$ as one of the $\frac{H}{32} \times \frac{W}{32} $ feature vectors of $\mathbf{X}_i = f_{\theta_{f}}(\mathbf{x}_i)$, and $\mathcal{P}_{\mathbf{y}_i} \subset \mathcal{P}$ is the set of prototypes with class label $\mathbf{y}_i$. 
For each training image, the cluster loss in~\eqref{eq:ell_CT} encourages the input image to have at least one local feature close to one of the prototypes of its own class, while the separation loss in~\eqref{eq:ell_SP} ensures all local features to be far from the prototypes that are not from the image's class.

Note in~\eqref{eq:ell_PPN} that compared with the original ProtoPNet~\cite{chen2019looks}, we introduce the hinge loss~\cite{xing2018robust} on $\ell_{SP}(\cdot)$ to impose a margin that mitigates the risk of overfitting.
After each training epoch, we update each prototype $\mathbf{p}_m$ to be represented by the nearest latent feature vector $\mathbf{z}$ from all training images of the same class. 
Specifically,  we replace $\mathbf{p}_m$ with the nearest feature vector $\mathbf{z}$, as in:
\begin{equation}
\ \mathbf{p}_m \leftarrow \arg\min_{\mathbf{z} \in \mathbf{X}_{i\in\{1,...,|\mathcal{D}|\}}}||\mathbf{z} - \mathbf{p}_m||_2^2.
\label{eq:prototype_update}
\end{equation}

One practical limitation in \cite{chen2019looks} is that there is no guarantee of diversity among prototypes. Here, we enforce prototype diversity in $\mathcal{P}$ by ensuring that we never have the same training image used for updating more than one prototype in~\eqref{eq:prototype_update}. 
This is achieved with the following 2-step algorithm: 1) for each prototype $\mathbf{p}_m$, compute the distances to all training images of the same class, and sort the distances in ascending order; and 
2) select prototypes, where for the first prototype $\mathbf{p}_1$, we choose its nearest image and record the image index indicating that the image has been used, then for the second prototype $\mathbf{p}_2$, we do the same operation, but if the selected image has been used by previous prototypes (e.g., $\mathbf{p}_1$), we will skip this image and use one of the next nearest images to $\mathbf{p}_2$. The prototype selection stage is performed sequentially until all prototypes are updated by different training images. 
We show in the experiments that this greedy prototype selection strategy improves the prototype diversity of  ProtoPNet.

\subsubsection{Testing} The final prediction of BRAIxProtoPNet++ is obtained by averaging the GlobalNet and ProtoPNet predictions $\tilde{\mathbf{y}}^{G}$ and $\tilde{\mathbf{y}}^{L}$, and the interpretability is reached by showing the prototypes $\mathbf{p}_m \in \mathcal{P}$ that produced the largest max-pooling score, together with the corresponding similarity map $\mathbf{S}_{m}$.

\section{Experimental Results}

\subsection{Dataset}

The experiments are performed on a private large-scale breast screening mammogram ADMANI (Annotated Digital Mammograms and Associated Non-Image data) dataset. 
It contains high-resolution (size of 5416 $\times$ 4040) 4-view mammograms (L-CC, L-MLO, R-CC, and R-MLO) with diagnosis outcome per view (i.e., cancer and no cancer findings).
The dataset has 20592 (3262 cancer, 17330 non-cancer) training images and 22525 (806 cancer, 21719 non-cancer) test images, where there is no overlap of patient data between training and test sets. In the test set, 410 cancer images have lesion annotations labelled by experienced radiologists for evaluating cancer localisation. 
We also use the public Chinese Mammography Database (CMMD) ~\cite{cai2019breast} to validate the generalisation performance of BRAIxProtoPNet++.  
CMMD consists of 5200 (2632 cancer, 2568 non-cancer) 4-view test mammograms with size 2294 $\times$ 1914 pixels. 

\subsection{Experimental Setup}
The BRAIxProtoPNet++ is implemented in Pytorch~\cite{paszke2019pytorch}. The model is trained using Adam optimiser with an initial learning rate of 0.001, weight decay of 0.00001, and batch size of 16. The hyper-parameters in~\eqref{eq:ell_PPN_PP}--\eqref{eq:ell_PPN} are set using simple general rules (e.g., small values for $\omega, \lambda_1, \lambda_2$, $\alpha,\beta$ should be close to 1, and $\gamma >> 1$), but model results are relatively robust to a large range of their values (for the experiments below, we have: $\alpha=1, \beta=0.5, \omega=0.2,\lambda_1=0.1, \lambda_2=0.1, \gamma=10 $). 
For the two datasets, images are pre-processed using the Otsu threshold algorithm to crop the breast region, which is subsequently resized to $H = 1536, W = 768$.
The feature size $D = 128$ in~\eqref{eq:ell_SP}, and the temperature parameter $T=128$. 
The number of prototypes $M=400$ (200 for cancer class and 200 for non-cancer class).
We use EfficientNet-B0 \cite{tan2019efficientnet} and DenseNet-121~\cite{huang2017densely}
as SOTA backbones.
The training of BRAIxProtoPNet++ is divided into three stages: 1) training of backbone and GlobalNet, 2) training of ProtoPNet with a frozen backbone and GlobalNet, and 3) fine-tuning of the whole framework. 
Data augmentation techniques (e.g., translation, rotation, and scaling) are used to improve generalisation. 
All experiments are conducted on a machine with AMD Ryzen 9 3900X CPU, 2 GeForce RTX 3090 GPUs, and 32 GB RAM. The training of BRAIxProtoPNet++ takes about 28 hours, and the average testing time is about 0.0013 second per image. 

Classification accuracy is assessed with the area under the receiver operating characteristic curve (AUC).
To evaluate model interpretability, we measure the area under the precision recall curve (PR-AUC) for the cancer localisation on test samples. To evaluate prototype diversity, we calculate the mean pairwise cosine distance and $L2$ distance between  learned prototypes from the same class.

\subsection{Results}

We compare our proposed method with the following models: EfficientNet-B0~\cite{tan2019efficientnet}, DenseNet-121~\cite{huang2017densely}, Sparse MIL~\cite{zhu2017deep}, GMIC~\cite{shen2021interpretable}, and ProtoPNet~\cite{chen2019looks}.
For all these models, we use the publicly available codes provided by the papers.
EfficientNet-B0 and DenseNet-121 are non-interpretable models. Sparse MIL can localise lesions by dividing a mammogram into regions that are classified using multiple-instance learning with a sparsity constraint. 
For a fair comparison, we use EfficientNet-B0 as the backbone for Sparse MIL.
GMIC uses a global module to select the most informative regions of an input mammogram, then it relies on a local module to analyse those selected regions, it finally employs a fusion module to aggregate the global and local outputs for classification. All methods above, and our BRAIxProtoPNet++, are trained on the training set from ADMANI, and tested on the ADMANI testing set and the whole CMMD dataset. 

\begin{table}
\centering
\caption{\small AUC results on ADMANI and CMMD datasets. The best result is in bold.
}
 \label{tab:ClassificationResults}
\resizebox{0.70\textwidth}{!}{
\arrayrulecolor{black}
\begin{tabular}{c|c|c|ccllllllllllll} 
\cline{1-5}\arrayrulecolor{black}\cline{6-6}
\multicolumn{3}{c|}{\multirow{2}{*}{Methods}}                                                                          & \multicolumn{2}{c}{Test AUC}                &  &  &  &  &  &  &  &  &  &  &  &   \\
\multicolumn{3}{c|}{}                                                                                                  & ~ ADMANI~~             & ~ CMMD~~             &  &  &  &  &  &  &  &  &  &  &  &   \\ 
\arrayrulecolor{black}\cline{1-5}
\multicolumn{3}{c|}{DenseNet-121 \cite{huang2017densely}}                                                                                      & 88.54                & 82.38                   &  &  &  &  &  &  &  &  &  &  &  &   \\
\multicolumn{3}{c|}{EfficientNet-B0 \cite{tan2019efficientnet}}                                                                                   & 89.62                & 76.41                &  &  &  &  &  &  &  &  &  &  &  &   \\
\multicolumn{3}{c|}{Sparse MIL \cite{zhu2017deep}}                                                                                        & 89.75                & 81.33                &  &  &  &  &  &  &  &  &  &  &  &   \\
\multicolumn{3}{c|}{GMIC \cite{shen2021interpretable}}                                                                                              & 89.98                & 81.03                &  &  &  &  &  &  &  &  &  &  &  &   \\
\multicolumn{3}{c|}{ProtoPNet (DenseNet-121) \cite{chen2019looks}}                                                                          & 87.12                & 80.23                &  &  &  &  &  &  &  &  &  &  &  &   \\
\multicolumn{3}{c|}{ProtoPNet (EfficientNet-B0) \cite{chen2019looks}}                                                                       & 88.30                & 79.61                &  &  &  &  &  &  &  &  &  &  &  &   \\ 
\cline{1-5}
\multirow{6}{*}{Ours (DenseNet-121)}    & \multirow{3}{*}{w/o KD}                     & ProtoPNet                      &  87.32              &  80.09                    &  &  &  &  &  &  &  &  &  &  &  &   \\
                                        &                                             & GlobalNet                      &  88.45             &   82.42                   &  &  &  &  &  &  &  &  &  &  &  &   \\
                                        &                                             & Ensemble                       &  88.87              &   82.50                   &  &  &  &  &  &  &  &  &  &  &  &   \\ 
\cline{2-5}
                                        & \multicolumn{1}{c|}{\multirow{3}{*}{w/ KD}} & \multicolumn{1}{c|}{ProtoPNet} & 88.35                &  80.67
                    &  &  &  &  &  &  &  &  &  &  &  &   \\ 
                                        & \multicolumn{1}{l|}{}                       & \multicolumn{1}{c|}{GlobalNet} & 88.61                &  82.52
                    &  &  &  &  &  &  &  &  &  &  &  &   \\ 
                                        & \multicolumn{1}{l|}{}                       & \multicolumn{1}{c|}{Ensemble}  & 89.54                & \textbf{82.65}             &  &  &  &  &  &  &  &  &  &  &  &   \\ 
\cline{1-5}
\multirow{6}{*}{Ours (EfficientNet-B0)} & \multirow{3}{*}{w/o KD}                     & ProtoPNet                      & 88.63                & 79.01                &  &  &  &  &  &  &  &  &  &  &  &   \\
                                        &                                             & GlobalNet                      & 90.11                & 76.50                &  &  &  &  &  &  &  &  &  &  &  &   \\
                                        &                                             & Ensemble                       & 90.18                & 80.45                &  &  &  &  &  &  &  &  &  &  &  &   \\ 
\cline{2-5}
                                        & \multirow{3}{*}{w/ KD}                      & ProtoPNet                      & 89.55                & 79.86                &  &  &  &  &  &  &  &  &  &  &  &   \\
                                        &                                             & GlobalNet                      & 90.12                & 76.47                &  &  &  &  &  &  &  &  &  &  &  &   \\
                                        &                                             & Ensemble                       & \textbf{ 90.68 }     & 81.65      &  &  &  &  &  &  &  &  &  &  &  &   \\
\cline{1-5}
\end{tabular}}
\vspace{-2pt}
\end{table}

Table~\ref{tab:ClassificationResults} shows the test AUC results of all methods on ADMANI and CMMD datasets. 
For our BRAIxProtoPNet++, we present the classification results of the ProtoPNet and GlobalNet branches independently, and their ensemble result to show the importance of combining the classification results of both branches. We also show results
with (w/KD) and without (w/o KD) distilling the knowledge from GlobalNet to train the ProtoPNet branch. Our best result is achieved with the ensemble model trained with KD, which reaches SOTA results on ADMANI and CMMD datasets.
Note that original ProtoPNet's AUC~\cite{chen2019looks} is worse than the non-interpretable global classifers EfficientNet-B0 and DenseNet-121. However, the application of KD to the original  ProtoPNet provides substantial AUC improvement, showing the importance of KD. 
It is observed that using DenseNet-121 as backbone exhibits better generalisation results on CMMD than using EfficientNet-B0, which means that DenseNet-121 is more robust against domain shift \cite{he2020sample}.
For the GMIC model, we note a discrepancy on the CMMD result on Table~\ref{tab:ClassificationResults} (AUC=81.03) and the published result in~\cite{stadnick2021meta} (AUC=82.50). This is explained by the different training set and input image setup used by GMIC in~\cite{stadnick2021meta}, so to enable a fair comparison, we present the result by GMIC with the same experimental conditions as all other methods in the Table.

\begin{figure}[t!]
    \centering
    \includegraphics[width=0.99\textwidth]{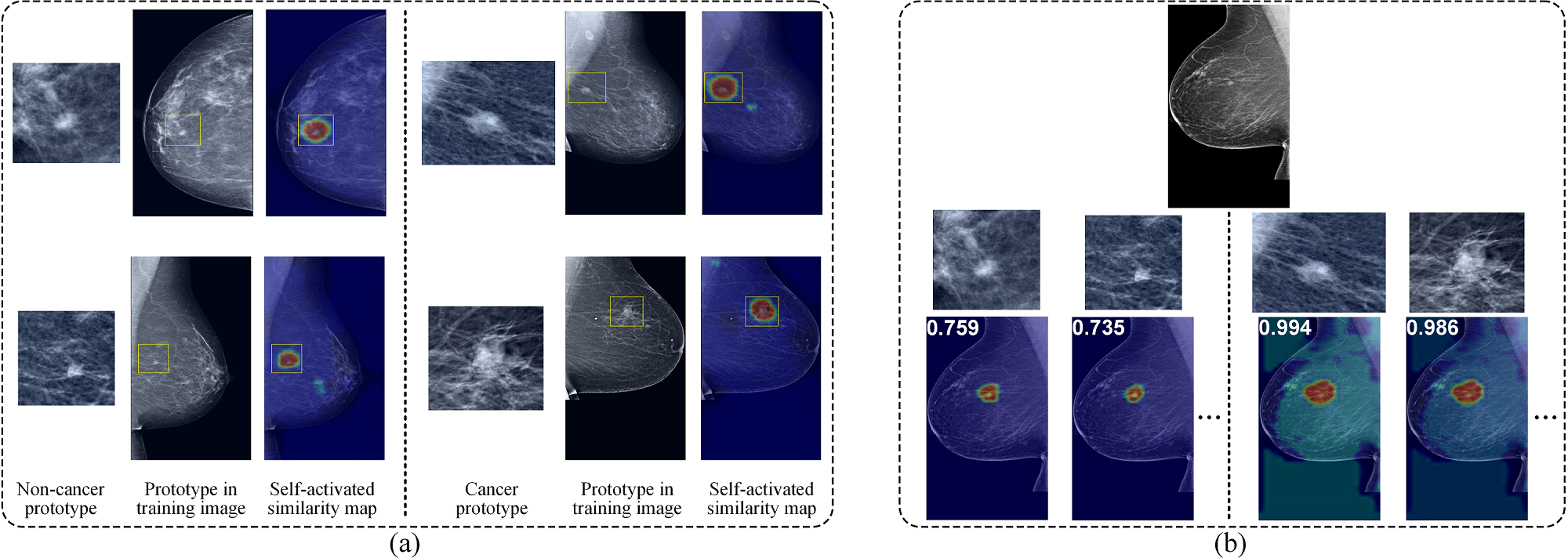}
    \vspace{-8pt}
    \caption{(a) Examples of non-cancer (left) and cancer prototypes (right) from BRAIxProtoPNet++. (b) The interpretable classification. First row: test image. Second row: top-2 activated non-cancer (left) and cancer (right) prototypes. Third row: similarity maps with the max-pooling score for classification. 
    }
    \label{fig:prototypes and reasoning}
    \vspace{-12pt}
\end{figure}

Fig.~\ref{fig:prototypes and reasoning} (a) displays the learned non-cancer and cancer prototypes and their source training images. 
We can see that the cancer prototypes come from regions containing cancerous visual biomarkers (e.g., malignant mass) which align with radiologists' criterion for breast cancer diagnosis, while non-cancer prototypes are from normal breast tissues or benign regions.
Fig.~\ref{fig:prototypes and reasoning} (b) shows the interpretable reasoning of BRAIxProtoPNet++ on a cancerous test image. We can see that our model classifies the image as cancer because the lesion present in the image looks more like the cancer prototypes than the non-cancer ones. 

\begin{figure}
\vspace{-12pt}
\begin{minipage}[t]{0.294\linewidth}
    \centering
    \captionsetup{width=.95\linewidth}
    \includegraphics[width=1.0\textwidth]{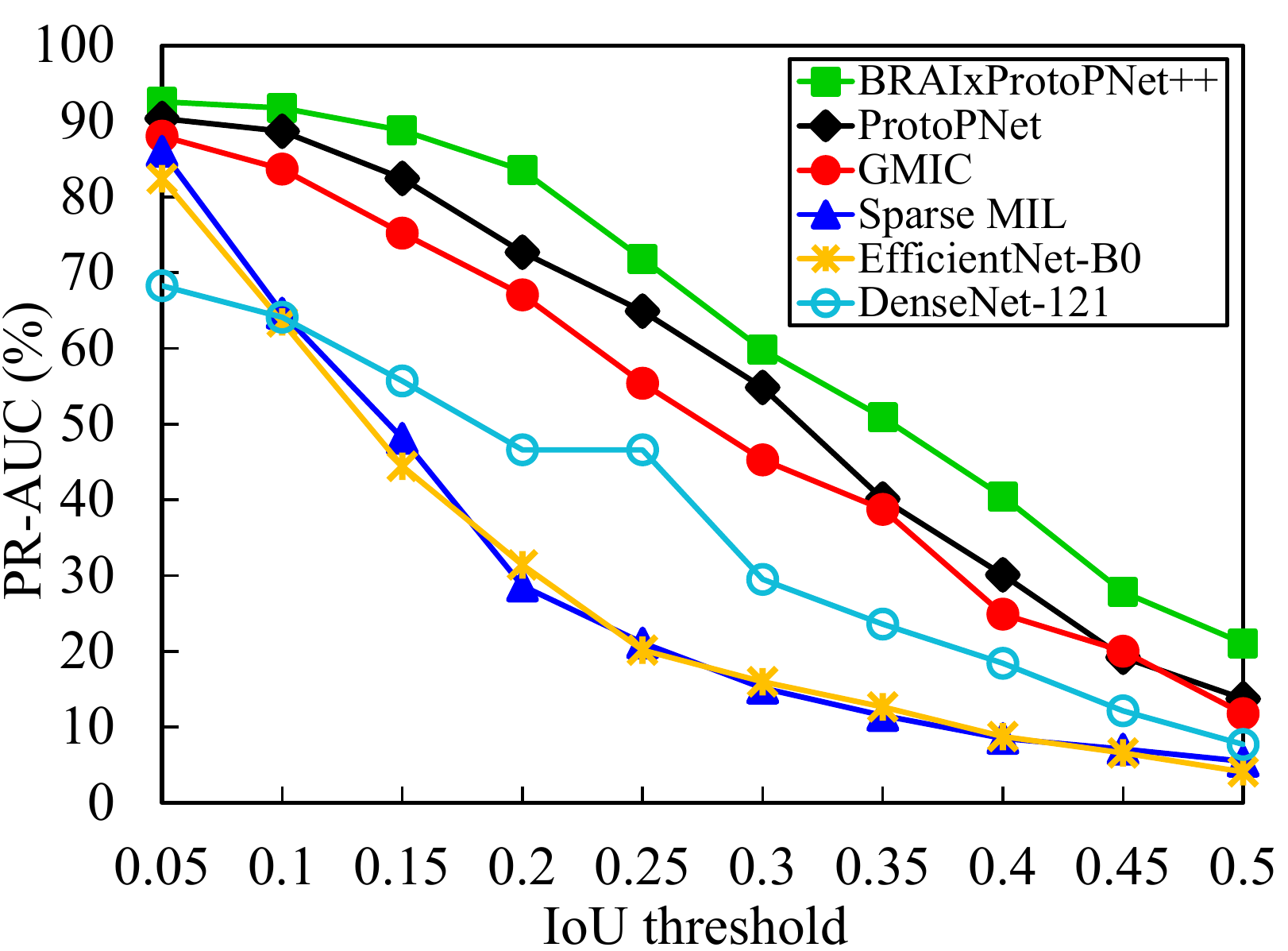}
    \vspace{-15pt}
    \caption{PR-AUC in different IoU thresholds.}
    \label{fig:PR-AUC curves}
\end{minipage}%
\begin{minipage}[t]{0.705\linewidth}
    \centering
    \captionsetup{width=.92\linewidth}
    \includegraphics[width=1.0\textwidth]{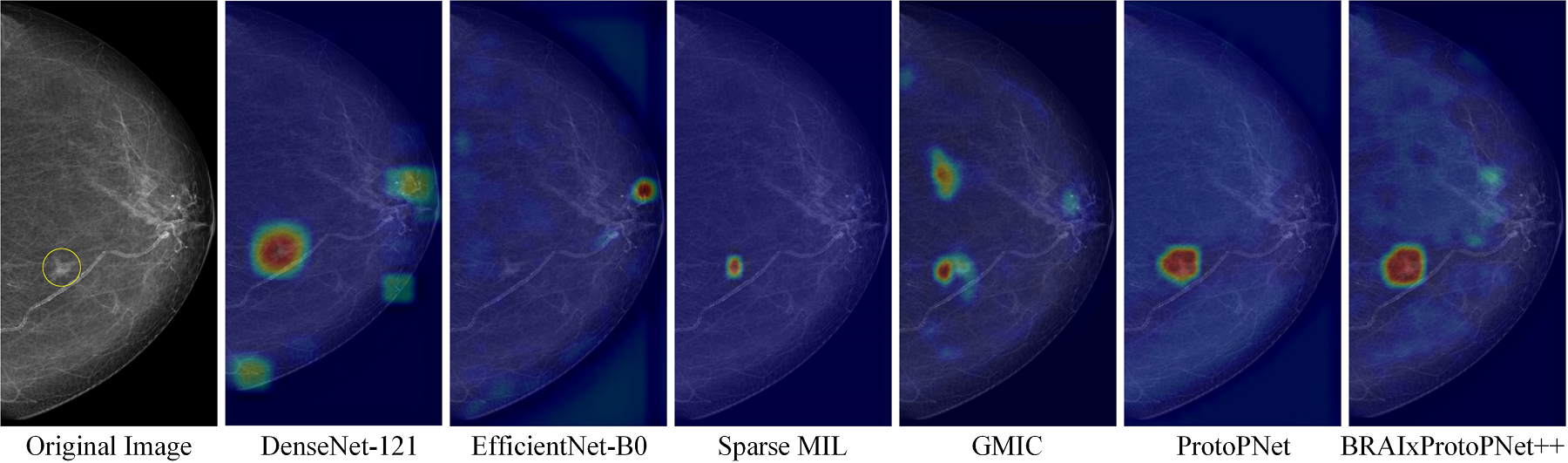}
    \vspace{-15pt}
    \caption{Visual results of cancer localisation. Yellow circle in the original image indicates cancer region.}
\label{fig:visualcomparison}
\end{minipage}
\vspace{-15pt}
\end{figure}

We also evaluate model interpretability by assessing the cancer localisation. 
The cancer regions are predicted by applying a threshold of 0.5 on the Grad-CAM (EfficientNet-B0 and DenseNet-121), malignant map (Sparse MIL), salience map (GMIC), and similarity map with the top-1 activated cancer prototype (ProtoPNet and BRAIxProtoPNet++). 
For all models, we exclude images with classification probability less than 0.1 since they are classified as non-cancer. 
When computing PR-AUC, we threshold several values (from 0.05 to 0.5) for the intersection over union (IoU) between predicted cancer region and ground-truth cancer mask to obtain a series of PR-AUC values, as shown in Fig.~\ref{fig:PR-AUC curves}. We can see that our BRAIxProtoPNet++ consistently achieves superior cancer localisation performance over the other methods  under different IoU thresholds. Fig.~\ref{fig:visualcomparison} displays a visual comparison of cancer localisation, where we can observe that the prototype-based methods can more accurately detect the cancer region.

We also investigate the effect of our proposed  prototype selection strategy on the prototype diversity and classification accuracy. Table~\ref{tab:prototypediversity} shows that our selection strategy can significantly increase prototype diversity (note larger cosine and $L2$ distances), which is beneficial for interpretability and classification.  
\begin{table}
\vspace{-20pt}
\caption{The effect of greedy prototype selection strategy on ADMANI dataset.}
 \label{tab:prototypediversity}
\centering
\begin{tabular}{llllll} 
\hline
\multicolumn{1}{c|}{\multirow{2}{*}{Methods}}        & \multicolumn{2}{c|}{Cosine distance}                             & \multicolumn{2}{c|}{$L2$ distance}                                & \multicolumn{1}{c}{\multirow{2}{*}{~AUC~}}  \\ 
\cline{2-5}
\multicolumn{1}{c|}{}                                & \multicolumn{1}{c}{Non-cancer} & \multicolumn{1}{c|}{~ Cancer~~} & \multicolumn{1}{c}{Non-cancer} & \multicolumn{1}{c|}{~ Cancer~~} & \multicolumn{1}{c}{}                        \\ 
\hline
\multicolumn{1}{c|}{ProtoPNet w/o greedy selection} & \multicolumn{1}{c}{0.034}      & \multicolumn{1}{c|}{0.061}      & \multicolumn{1}{c}{0.805}     & \multicolumn{1}{c|}{0.827}      & \multicolumn{1}{c}{88.11}                   \\
\multicolumn{1}{c|}{ProtoPNet w/ greedy selection}  & \multicolumn{1}{c}{0.074}      & \multicolumn{1}{c|}{0.094}      & \multicolumn{1}{c}{1.215}      & \multicolumn{1}{c|}{1.712}      & \multicolumn{1}{c}{88.30}                   \\ 
\hline
                                                     &                                &                                 &                                &                                 &                                             \\
                                                     &                                &                                 &                                &                                 &                                             \\
                                                     &                                &                                 &                                &                                 &                                             \\
                                                     &                                &                                 &                                &                                 &                                             \\
                                                     &                                &                                 &                                &                                 &                                            
\end{tabular}
\vspace{-72pt}
\end{table}

\section{Conclusion}
In this paper, we presented BRAIxProtoPNet++ to realise accurate mammogram classification with effective prototype-based interpretability.
Our approach has been designed to enable the integration of prototype-based interpretable model to any highly accurate global mammogram classifier, where we distill the knowledge of the global model when training the prototype-based model to increase the classification accuracy of the ensemble.
We also proposed a method to increase the diversity of the learned prototypes.
Experimental results on private and public datasets show that BRAIxProtoPNet++ has SOTA classification and interpretability results.
One potential limitation in our BRAIxProtoPNet++ is that the learned prototypes are class-specific, we will explore to learn class-agnostic prototypes for mammogram classification in the future work. 

\subsubsection{Acknowledgement.} This work was supported by funding from the Australian Government under the Medical Research Future Fund - Grant MRFAI000090 for the Transforming Breast Cancer Screening with Artificial Intelligence (BRAIx) Project, and the Australian Research Council through grants DP180103232 and FT190100525.

%
%
%
\bibliographystyle{splncs04}
\bibliography{mybibliography}
%




\end{document}